\documentclass[letterpaper]{article} % DO NOT CHANGE THIS
\usepackage{aaai25}  % DO NOT CHANGE THIS
\usepackage{times}  % DO NOT CHANGE THIS
\usepackage{helvet}  % DO NOT CHANGE THIS
\usepackage{courier}  % DO NOT CHANGE THIS
\usepackage[hyphens]{url}  % DO NOT CHANGE THIS
\usepackage{graphicx} % DO NOT CHANGE THIS
\usepackage{natbib}  % DO NOT CHANGE THIS AND DO NOT ADD ANY OPTIONS TO IT
\usepackage{caption} % DO NOT CHANGE THIS AND DO NOT ADD ANY OPTIONS TO IT
\usepackage{algorithm}
\usepackage{algorithmic}

\usepackage{subfig}
\usepackage{multirow}
\usepackage{amsmath}
\usepackage[table,xcdraw]{xcolor}
\usepackage{amssymb}
\usepackage{newfloat}
\usepackage{listings}
\DeclareCaptionStyle{ruled}{labelfont=normalfont,labelsep=colon,strut=off} % DO NOT CHANGE THIS
\floatstyle{ruled}
\newfloat{listing}{tb}{lst}{}
\floatname{listing}{Listing}
\title{RaCMC: Residual-Aware Compensation Network with Multi-Granularity Constraints for Fake News Detection}
\author {
    Xinquan Yu\textsuperscript{\rm 1},
    Ziqi Sheng\textsuperscript{\rm 1},
    Wei Lu\textsuperscript{\rm 1}\thanks{Corresponding authors},
    Xiangyang Luo\textsuperscript{\rm 2}\footnotemark[1],
    Jiantao Zhou\textsuperscript{\rm 3}
}
\affiliations {
    \textsuperscript{\rm 1}School of Computer Science and Engineering, Ministry of Education Key Laboratory of Information Technology, Guangdong Province Key Laboratory of Information Security Technology, Sun Yat-sen University, Guangzhou 510006, China\\
    \textsuperscript{\rm 2}State Key Laboratory of Mathematical Engineering and Advanced Computing, Zhengzhou 450002, China\\
    \textsuperscript{\rm 3}State Key Laboratory of Internet of Things for Smart City and the Department of Computer and Information Science, Faculty of Science and Technology, University of Macau, Macau 999078, China\\
    yuxq28@mail2.sysu.edu.cn, shengzq@mail2.sysu.edu.cn, luwei3@mail.sysu.edu.cn, luoxy\_ieu@sina.com, jtzhou@um.edu.mo
}

\begin{document}

\maketitle

\begin{abstract}
Multimodal fake news detection aims to automatically identify real or fake news, thereby mitigating the adverse effects caused by such misinformation. Although prevailing approaches have demonstrated their effectiveness, challenges persist in cross-modal feature fusion and refinement for classification. 
To address this, we present a residual-aware compensation network with multi-granularity constraints (RaCMC) for fake news detection, that aims to sufficiently interact and fuse cross-modal features while amplifying the differences between real and fake news.
First, a multiscale residual-aware compensation module is designed to interact and fuse features at different scales, and ensure both the consistency and exclusivity of feature interaction, thus acquiring high-quality features.
Second, a multi-granularity constraints module is implemented to limit the distribution of both the news overall and the image-text pairs within the news, thus amplifying the differences between real and fake news at the news and feature levels.
Finally, a dominant feature fusion reasoning module is developed to comprehensively evaluate news authenticity from the perspectives of both consistency and inconsistency. Experiments on three public datasets, including Weibo17, Politifact and GossipCop, reveal the superiority of the proposed method.

\end{abstract}

\section{Introduction}

\label{sec:intro}
The rapid development of social networks accelerates self-media, enabling ordinary individuals to become publishers of daily news. This 
%widens the horizons of internet users and 
accelerates the speed at which people receive information. However, it also causes the widespread dissemination of misinformation due to out-of-context interpretations, exaggerations and malicious falsifications by publishers, which seriously endangers social security and stability \cite{lazer2018science,fleming2020coronavirus}. Although national authorities and social platforms establishing news verification departments\footnote{https://www.piyao.org.cn/index.htm}\footnote{https://www.politifact.com/} %\footnote{https://www.dhs.gov/}
to distinguish between real and fake news, the massive amount of news places a significant burden on news verifiers. Consequently, automatic fake news detection becomes a research hotspot.

Early methods were designed for pure text news, aiming to fully explore all information within the text. For example,  
DSTS \cite{ma2015detect} designed a dynamic series-time structure, aimed at capturing features related to temporal variations during news dissemination. 
HiMaP \cite{mishra2020fake} designed a high-order mutual-attention progression method to capture real and fake news propagation patterns. 

\begin{figure}[t]
	\centering
	{\includegraphics[width=3.3in]{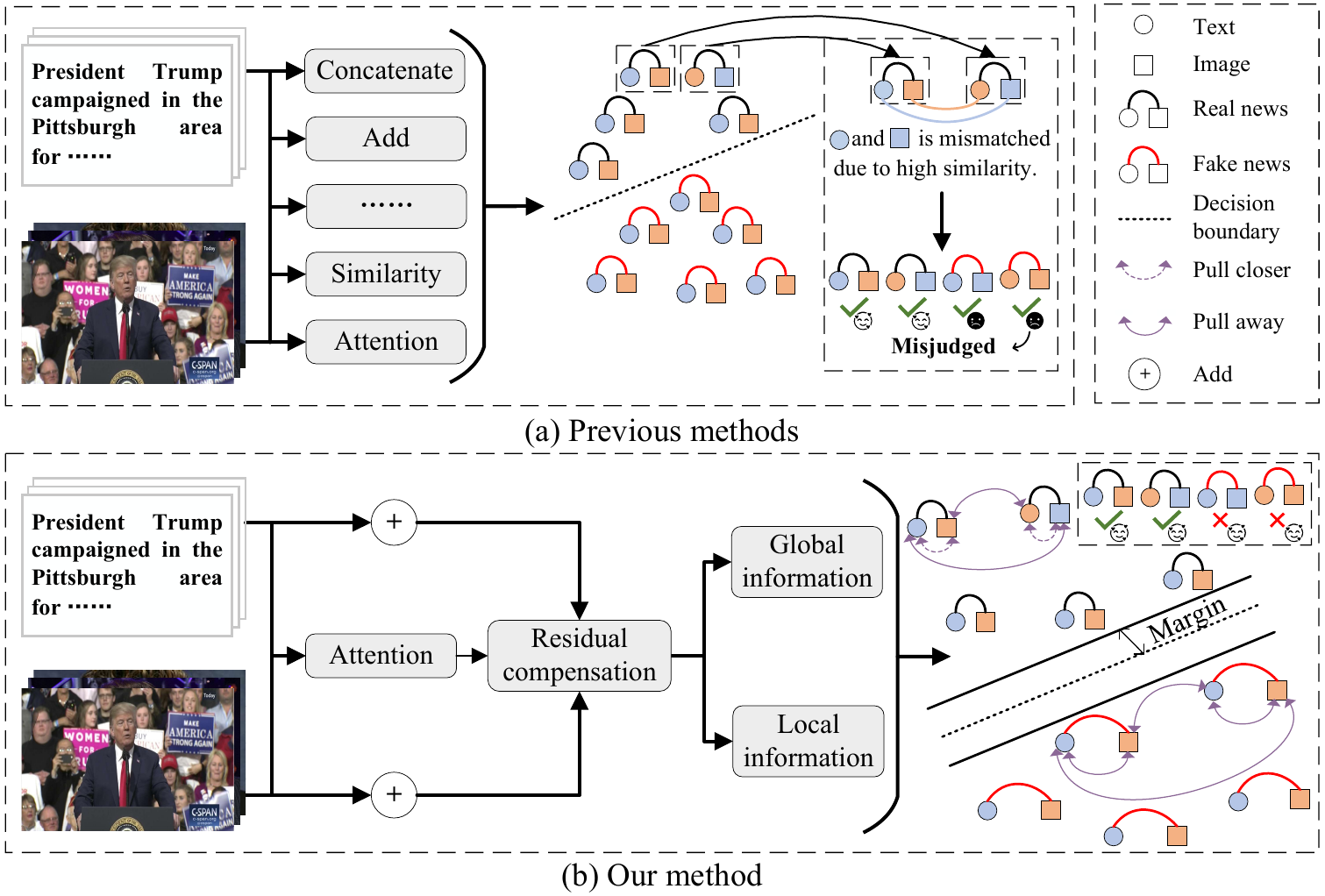}}
	\caption{Comparison of the fusion and reasoning processes. (a) Previous methods exhibit inadequate cross-modal feature fusion and may misjudge real news with high similarity. (b) Our method refines cross-modal fusion with residual compensation and multi-scale information mining. Besides, it amplifies the differences between real and fake news in terms of news and feature levels.}
	\label{fig0}
\end{figure}

With the development of Internet, news evolved from pure text to image-text pairs, leading to the emergence of multimodal fake news detection. Such methods aim to fully explore the cross-modal feature alignment, interaction and fusion, thus identifying fake news.
For example, att-RNN \cite{jin2017multimodal} was the first to utilize image information to aid in detection. 
Although it was recognized early on that mining the correlation information between text and image could improve detection accuracy, it did not consider the complementary information of different modalities and the differences in representation space \cite{zhang2019multi,guan2021multimodal}. 
To this end, SAFE \cite{zhou2020similarity} fed the correlation between text and visual information into a classifier for detection.  
MCNN \cite{xue2021detecting} utilized the similarity of text and visual information. 
FND-CLIP \cite{zhou2023multimodal} utilized SE-Net \cite{hu2018squeeze} to reweight features from different modalities. 
CFFN \cite{li2023cross} mined the consistency and inconsistency of cross-modal features.
MRAN \cite{yang2024mran} utilized an attention mechanism for cross-modal interaction and later achieved fusion through concatenation.

Although existing methods have achieved a certain degree of performance, several defects remain as follows: 1) Inadequate cross-modal feature fusion. As shown in Figure \ref{fig0}(a), existing methods generally adopt concatenation, addition, similarity reweighting and attention to fuse features from different modalities. However, linear operations such as concatenation and addition do not consider contextual semantic information. 
Although similarity reweighting and attention can capture the consistency of cross-modal features, it represents only a part of the original information. 
2) Fuzzy feature classification. Existing methods achieve detection solely at the news level (image-text pairs), and do not consider the differences at the feature level (between image and text). As shown in Figure \ref{fig0}(a), this may result in incorrect pairing of image-text pairs due to the high similarity of cross-modal features from different news.

To address the above issues, we propose a Residual-aware Compensation framework with Multi-granularity Constraints (RaCMC). Specifically, to sufficiently fuse cross-modal features, we design a multiscale residual-aware compensation module. It first filters useless interference information and captures cross-modal consistency information utilizing an attention mechanism with a mask, and then fuses features from different sources and modalities by utilizing residual connections and multiscale pooling, so as to mine both complete unimodal features and multimodal features (as shown in Figure \ref{fig0}(b)). 
Next, to enable feature refinement classification, we develop a multi-granularity constraints module. It first amplifies the difference between real and fake news at news level by utilizing the maximum mean difference. Then, we further amplify the difference at feature level (as shown in Figure \ref{fig0}(b)). For real news, two operations of pulling close and pulling away are adopted to ensure that the features within the same image-text pair remain closely aligned, insulated from the influence of high-similarity features from other real news. For fake news, only the pulling away operation is adopted to sever the cohesion of all image-text pairs and keep them separated from each other.
Finally, we comprehensively reason about the news authenticity in terms of consistency and inconsistency. The main contributions of this paper are as follows:
\begin{enumerate}
	\item
	We propose RaCMC, a residual-aware compensation network with multi-granularity constraints for fake news detection, aiming to sufficiently fuse cross-modal features and amplify the difference between real and fake news.
	\item
	We design a multiscale residual-aware compensation module to interact and fuse the cross-modal multiscale features. Then, we present the interaction constraints to ensure the consistency and exclusivity of features after the interaction.
	\item
	We devise a multi-granularity constraints module that amplifies the distinctions between real and fake news at both the news and feature levels.
\end{enumerate}

\section{Related Works}
\label{sec:Related Works}
Existing multimodal fake news detection methods can be broadly divided into three categories based on the information utilized: image-text-based detection, social context-based detection and external knowledge-based detection. 

\textbf{Image-text-based detection}. Such methods aim to investigate the alignment and fusion of image and text modalities. It mainly includes: 
Spotfake \cite{singhal2019spotfake} first utilized the pre-trained VGG-19 and BERT model to extract the image and text features, they then concatenated the two features to identify real or fake news.
MVAE \cite{khattar2019mvae} utilized the Encoder-Decoder network to model shared representation between text and image.
HMCAN \cite{qian2021hierarchical} utilized cross-attention to interact cross-modal features and then performed feature fusion through concatenation.
BMR \cite{ying2023bootstrapping} designed a multi-gate fusion expert network for feature fusion.
QMFND \cite{qu2024qmfnd} encoded multimodal features into variational quantum circuits to achieve better accuracy.
FSRU \cite{lao2024frequency} made the first attempt to utilize spectral features for detection.

\textbf{Social context-based detection}. Such methods aim to model the news propagation process and utilize poster information, posting history, comments, etc., to aid in detection. It mainly includes: 
UPFD \cite{dou2021user} analyzed user preferences such as posting history.
HCCIN \cite{wu2023human} interacted comment information with image and text information through an attention mechanism.
MRHFR \cite{wu2023see} designed an explored coherence constraint reasoning layer to infer the coherence between comments and news, and evaluated the semantic deviation between unimodal and multimodal features.

\textbf{External knowledge-based detection}. Such methods aim to utilize external news information such as entity enhancement, retrieval, to aid in detection. It mainly includes: 
EM-FEND \cite{qi2021improving} extracted visual entities and the embedded text in images to model the high-level visual semantics.
HSEN \cite{zhang2023hierarchical} utilized textual entities as prompt subject vocabulary to generate image captions, thereby enriching image knowledge information.
Lin et al. presented a 
dataset named CFEVER \cite{lin2024cfever}, which utilizes the evidence sentences sourced from single or multiple pages in Chinese Wikipedia.

In addition, there are early fake news detection methods \cite{wang2018eann,zhang2020bdann,fang2024nsep}, tamper-specific fake news detection \cite{shao2023detecting,shao2024detecting,li2024towards}, 
unsupervised detection \cite{wang2023positive,yin2024gamc}, etc.

In this paper, we still focus on image-text-based detection. Addressing the problems mentioned above, 
we construct a residual-aware compensation network with multi-granularity constraints for fake news detection. 

\section{Methodology}
\label{sec:Methodology}
 
\begin{figure*}
	\centering
	{\includegraphics[width=\linewidth]{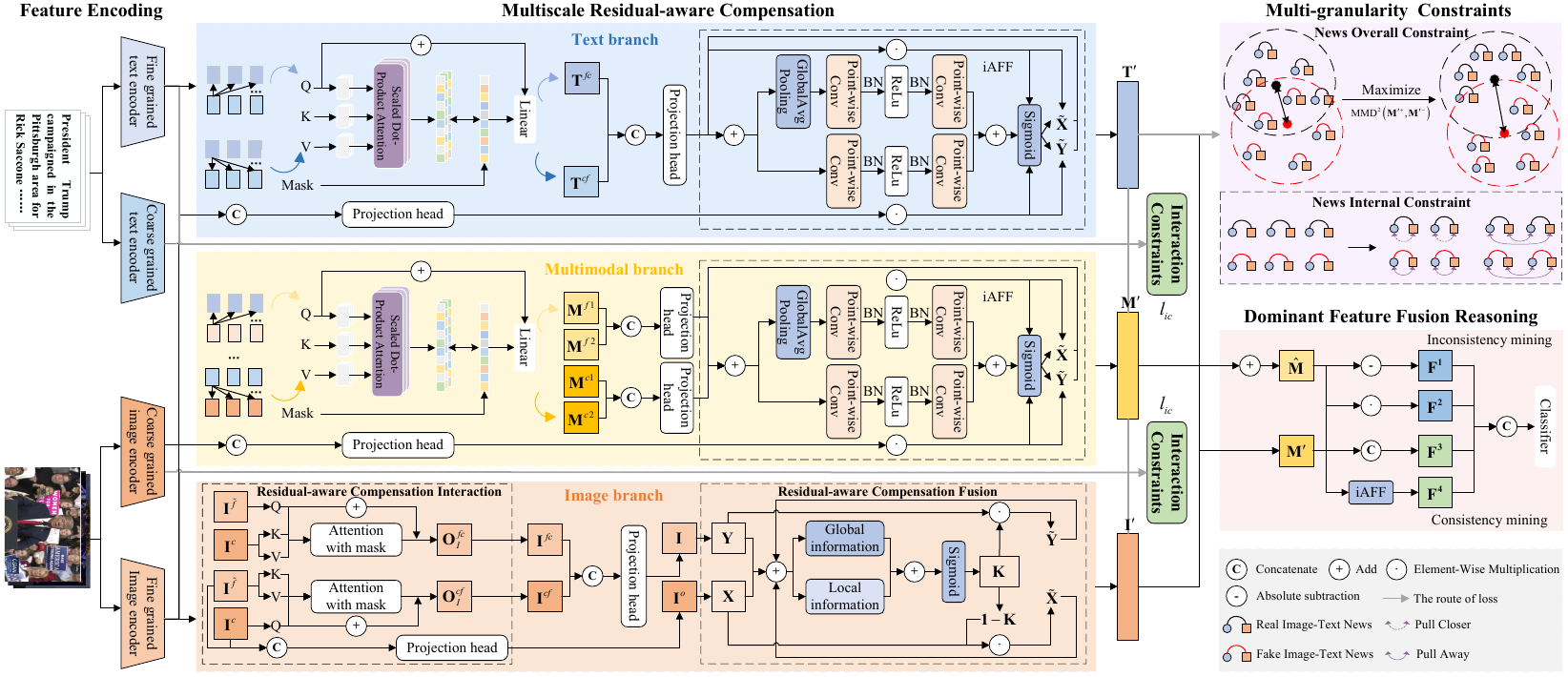}}
	\caption{The network architecture of RaCMC. It consists of four components: (1) Feature Encoding: this module extracts features with different granularities utilizing a coarse-fine dual-grained encoder. (2) Multiscale Residual-aware Compensation: this module sufficiently interacts and fuses features from different modalities and sources. It has three branches, processing text (blue block), multimodal (yellow block), and image (orange block) features from top to bottom. The exact execution process is detailed in the image branch. (3) Multi-granularity Constraints: this module amplifies the differences between real and fake news at both the news and feature levels. (4) Dominant Feature Fusion Reasoning: this module evaluates news authenticity in terms of both consistency and inconsistency.}
	\label{fig1}
\end{figure*}

As illustrated in Figure \ref{fig1}, RaCMC comprises four components: feature encoding, multiscale residual-aware compensation (MRC), multi-granularity constraints (MGC), and dominant feature fusion reasoning (DFR). Next, we explain each component in detail.

\subsection{Feature Encoding}
\label{subsec:Feature Encoding}
Multimodal news consists of an image-text pair. To mine the features of each modality in more detail and retrieve usable information, we adopt a coarse-fine dual-grained encoder to obtain the original feature representations.

For the text modality, we first utilize the pre-trained BERT model to capture subtle contextual differences, so as to extract the fine-grained text features denoted as $\mathbf{T}^{f} \in \mathbb{R}^{B \times n_1}$, where $B$ denotes batch size.
Then, we utilize the pre-trained CLIP model to acquire joint information between text and image, so as to extract the coarse-grained text features denoted as $\mathbf{T}^{c} \in \mathbb{R}^{B \times N}$.
For the image modality, a similar design is adopted. Specifically, we utilize the pre-trained ResNet-101 model and the pre-trained CLIP model to extract deep features and understand semantic relationships of the image-text pair, respectively. Then, we can obtain $\mathbf{I}^{f} \in \mathbb{R}^{B \times n_2}$ and $\mathbf{I}^{c} \in \mathbb{R}^{B \times N}$.  
Finally, the dimensions of all features mentioned above are projected to ${B \times N}$ for subsequent interaction, i.e., $\mathbf{T}^{\tilde{f}}, \mathbf{T}^{c}, \mathbf{I}^{\tilde{f}}, \mathbf{I}^{c} \in \mathbb{R}^{B \times N}$.

\subsection{Multiscale Residual-aware Compensation}
\label{subsec:Multiscale Residual Perception Compensation}

To better utilize the features extracted above, we present MRC, which consists of three layers: residual-aware compensation interaction, residual-aware compensation fusion and interaction constraints.

\subsubsection{Residual-aware Compensation Interaction}
\label{subsubsec:MultiscaleInteractionPerception} 
We propose utilizing an attention mechanism to achieve interaction perception of coarse-fine granularity features, thereby compensating for existing granularity features based on the information from another granularity. Specifically, we first execute the scaled dot-product attention,
\begin{equation}
\label{eq1}
\mathrm{Attention} (\mathbf{Q} ,\mathbf{K} ,\mathbf{V} ) = \mathrm{softmax}  \left ( \frac{\mathbf{Q} \mathbf{K} ^t}{\sqrt{d}} \right ) \mathbf{V}
\end{equation}
where $\mathbf{Q}$, $\mathbf{K}$ and $\mathbf{V}$ denote the queries, keys and values, respectively. $\sqrt{d}$ denotes the scaling factor. 

Considering that the features are extracted from different encoders, direct feature interaction may be interfered with by noise or irrelevant features. We design a variable mask threshold matrix denoted as $\Theta$, computed by concatenating queries and keys into a fully connected network \cite{wang2023positive}. 
Based on this, we can calculate the mask matrix denoted as $\mathbf{\Omega}$,
\begin{equation}
\mathbf{\Omega}_{i,j} = \left\{\begin{matrix} 1,   &\mathrm{if} \;\; \mathbf{S}_{i,j} \ge \Theta_{i,j}
\\0, &\mathrm{otherwise}
\end{matrix}\right.
\end{equation}
where $\mathbf{S} = \mathrm{softmax}  \left ( \frac{\mathbf{Q} \mathbf{K} ^t}{\sqrt{d}} \right)$, denotes the similarity score matrix. In this way, Eq. \eqref{eq1} is converted into
\begin{equation}
\label{eq3}
\mathrm{Attention} (\mathbf{Q} ,\mathbf{K} ,\mathbf{V}, \mathbf{\Omega}) = \mathrm{softmax}  \left (\mathbf{\Omega} \times \frac{ \mathbf{Q} \mathbf{K} ^t}{\sqrt{d}} \right ) \mathbf{V}
\end{equation}

Then, we utilize the multi-head attention to enrich the feature representation. Here, we first project $\mathbf{Q}$, $\mathbf{K}$ and $\mathbf{V}$ to different feature subspaces, then execute Eq. \eqref{eq3} to them in parallel. This process can be formulated as
\begin{equation}
\mathrm{head} _i = \mathrm{Attention} (\mathbf{Q}\mathbf{W}_{i}^{q} ,\mathbf{K}\mathbf{W}_{i}^{k} ,\mathbf{V}\mathbf{W}_{i}^{v}, \mathbf{\Omega})
\end{equation}
and
\begin{equation}
\label{eq5}
\begin{split}
\mathbf{O}  &= \mathrm{Multihead} (\mathbf{Q} ,\mathbf{K} ,\mathbf{V}, \mathbf{\Omega}) \\
&= \mathrm{Concat}\left(\mathrm{head}_1,\cdots ,\mathrm{head}_h \right)\mathbf{W}^{o}
\end{split}
\end{equation}
where all $\mathbf{W}$ are learnable parameters. $\mathbf{O}$ denotes the output, $h$ denotes the number of heads.

Based on the above, we can achieve interaction perception of various modality features. Specifically, for text features $\mathbf{T}^{\tilde{f}}$ and $\mathbf{T}^{c}$, we first execute Eq. \eqref{eq5} with $\mathbf{T}^{\tilde{f}}$ as $\mathbf{Q}$, $\mathbf{T}^{c}$ as $\mathbf{K}$ and $\mathbf{V}$, we have
\begin{equation}
\mathbf{O}^{fc}_{T} = \mathrm{Multihead} (\mathbf{T}^{\tilde{f}} ,\mathbf{T}^{c},\mathbf{T}^{c}, \mathbf{\Omega}^{T^{\tilde{f}},T^c})
\end{equation}
where $\mathbf{O}^{fc}_{T} \in \mathbb{R}^{B \times N}$. Then, we further enrich the feature representation through a residual block,
\begin{equation}
\label{eq7}
\mathbf{T}^{fc} = \mathbf{T}^{\tilde{f}} + \mathbf{O}^{fc}_{T}
\end{equation}
where $\mathbf{T}^{fc} \in \mathbb{R}^{B \times N}$ denotes the fine-grained interaction text features. Then, we exchange the assignments of $\mathbf{Q}$ with those of $\mathbf{K}$ and $\mathbf{V}$, and execute Eq. \eqref{eq5} and Eq. \eqref{eq7} to obtain the corase-grained interaction text features denoted as $\mathbf{T}^{cf}$.

Second, for image features, a similar operation is conducted, and we can get $\mathbf{I}^{fc}$ and $\mathbf{I}^{cf}$. Finally, for multimodal features, we only allow cross-modal features of the same granularity to interactively infiltrate, i.e., we divide the features into two groups for interaction, $\mathbf{T}^{\tilde{f}}$ and $\mathbf{I}^{\tilde{f}}$, $\mathbf{T}^{c}$ and $\mathbf{I}^{c}$. Thus, we obtain four multimodal features $\mathbf{M}^{f1}$, $\mathbf{M}^{f2}$, $\mathbf{M}^{c1}$ and $\mathbf{M}^{c2}$. 
For example, $\mathbf{M}^{f1}$ is calculated by
\begin{equation}
\mathbf{M}^{f1} = \mathbf{T}^{\tilde{f}} + \mathrm{Multihead} (\mathbf{T}^{\tilde{f}} ,\mathbf{I}^{\tilde{f}},\mathbf{I}^{\tilde{f}}, \mathbf{\Omega}^{T^{\tilde{f}},I^{\tilde{f}}})
\end{equation}

\subsubsection{Residual-aware Compensation Fusion}
\label{subsubsec:Residual Perception Compensation}
Feature fusion remains a formidable challenge in fake news detection. Prior research has explored various strategies, including element-wise addition, concatenation, SE-Net \cite{hu2018squeeze} and other methods. While these methods have yielded some promising outcomes, they are often limited in their ability to capture the nuanced interplay of diverse features. For instance, element-wise addition and concatenation are inherently linear operations that fail to account for the rich contextual nuances. Although SE-Net introduces non-linearity into the fusion process, it falls short in capturing the full details of input features due to its lack of local context sensitivity.

In light of these limitations, we propose adopting the iterative attention feature fusion named iAFF \cite{dai2021attentional}, which offers a comprehensive fusion that considers both global and local perspectives.
Specifically, for text features, we first concatenate $\mathbf{T}^{f}$ and $\mathbf{T}^{c}$, then feed it into a projection head containing a two-layer fully connected network with BatchNorm layer, ReLU activation function and Dropout layer,
\begin{equation}
\label{eq8}
\mathbf{T}^{o} = \mathrm{ProjectionHead}\left(\mathrm{Concat}\left( \mathbf{T}^{f}, \mathbf{T}^{c}  \right)\right)
\end{equation}
where $\mathbf{T}^{o} \in \mathbb{R}^{B \times N}$. Second, a similar operation is conducted for interaction text features $\mathbf{T}^{fc}$ and $\mathbf{T}^{cf}$, we can get $\mathbf{T} \in \mathbb{R}^{B \times N}$,
\begin{equation}
\label{eq9}
\mathbf{T} = \mathrm{ProjectionHead}\left(\mathrm{Concat}\left( \mathbf{T}^{fc}, \mathbf{T}^{cf}  \right)\right)
\end{equation}
Finally, we feed $\mathbf{X} = \mathbf{T}^{o}$ and $\mathrm{Y} = \mathbf{T}$ into iAFF,
\begin{align}
\label{eq10}
\mathrm{iAFF} \left( \mathbf{X},\mathbf{Y} \right) &= \mathbf{X} \times \mathbf{\tilde{K}} + \mathbf{Y} \times \left ( \mathbf{1 - \tilde{K}} \right )  \\
\label{eq11}
\mathbf{\tilde{K}} &= \sigma\left( \mathrm{G}\left ( \mathbf{\tilde{X} + \tilde{Y}} \right) + \mathrm{{L}}\left ( \mathbf{\tilde{X} +\tilde{Y}} \right)   \right) \\
\label{eq12}
\mathbf{\tilde{X}} &= \mathbf{X} \times \mathbf{K}, \mathbf{\tilde{Y}} = \mathbf{Y} \times \left( \mathbf{1-K} \right) \\
\label{eq13}
\mathbf{K} &= \sigma\left( \mathrm{G}\left ( \mathbf{{X} + {Y}} \right) + \mathrm{{L}}\left ( \mathbf{{X} + {Y}} \right)   \right)
\end{align}
where $\sigma$ denotes the Sigmoid function, $\mathrm{G}$ and $\mathrm{L}$ denote the global and local channel attention block, respectively. 
$\mathrm{L}$ contains a five-tiered structure, designed to extract local information, while $\mathrm{G}$ incorporates an additional global pooling layer, designed to extract global information.
Based on this, we obtain the multiscale residual compensation text features, denoted as $\mathbf{{T}'}$. 

For image features, we can also get $\mathbf{{I}'}$ in the same way. For multimodal features, by feeding $\mathbf{T}^{f}$ and $\mathbf{I}^{f}$ to Eq. \eqref{eq8}, we obtain $\mathbf{M}^{o1}$, where $\mathbf{M}^{o1} \in \mathbb{R}^{B \times N}$. Then, we have $\mathbf{M}^{1}$ by feeding $\mathbf{M}^{f1}$ and $\mathbf{M}^{f2}$ into Eq. \eqref{eq9}.
Next, we can get fine-grained multimodal features $\mathbf{M^f}$ by feeding $\mathbf{X} = \mathbf{M}^{o1}$ and $\mathbf{Y} = \mathbf{M^{1}}$ into Eqs. \eqref{eq10}-\eqref{eq13}.
In the same way, the coarse-grained multimodal features denoted as $\mathbf{M^c}$ can be obtained. Finally, we can get $\mathbf{{M}'}$ by concatenating $\mathbf{M^f}$ and $\mathbf{M^{c}}$, where $\mathbf{{M}'} \in \mathbb{R}^{B \times N}$.

\subsubsection{Interaction Constraints}
\label{subsubsec:Feature Interaction Constraints}

To ensure the quality of the feature post-interaction and prevent each modality feature from being largely divergent \cite{wu2023mfir}, we strategically engineer two complementary interaction constraint strategies: consistency and exclusivity. 
The former aims to enhance the uniformity between unimodal and multimodal features, ensuring the features derived from unimodal interaction exhibit a degree of similarity to multimodal features, and vice versa. 
Conversely, the latter is dedicated to upholding the intrinsic distinctiveness of each modality, mandating that the features post-interaction preserve the unique information inherent to their original modality.

The consistency constraint strategy is based on the KL divergence to measure the similarity between two input features. 
For instance, for the text features post-interaction $\mathbf{T}^{fc}$ and $\mathbf{T}^{cf}$, our goal is to ensure that $\mathbf{T}^{fc}$ and $\mathbf{T}^{cf}$ are as close as possible to multimodal features $\mathbf{{M}'}$, thereby guaranteeing consistency between the text interaction features and the multimodal features. To achieve this, we perform the following operations,
\begin{equation}
l_{c}^{T} = \mathrm{D}_{KL} \left( \mathbf{{M}'} \parallel \mathbf{T}^{fc} \right) + \mathrm{D}_{KL} \left( \mathbf{{M}'} \parallel \mathbf{T}^{cf} \right)
\end{equation}
Then, for image features, we can also get $l_{c}^{I}$ in the same way. For multimodal features, we utilize the dominant unimodal feature (i.e., $\mathbf{Q}$) during interaction to maintain consistency with the unimodal features,
\begin{equation}
\begin{split}
l_{c}^{M} &= \mathrm{D}_{KL} \left( \mathbf{{T}'} \parallel \mathbf{M}^{f1} \right) + \mathrm{D}_{KL} \left( \mathbf{{T}'} \parallel \mathbf{M}^{c1} \right) \\
&+ \mathrm{D}_{KL} \left( \mathbf{{I}'} \parallel \mathbf{M}^{f2} \right)+ \mathrm{D}_{KL} \left( \mathbf{{I}'} \parallel \mathbf{M}^{c2} \right)
\end{split}
\end{equation}

The exclusivity constraint strategy is based on the Frobenius norm to mine the intrinsic distinctiveness of features, avoiding the loss of their original unique properties after applying consistency constraints.
Here, we still take the text features as an example,
\begin{equation}
\begin{split}
l_{e}^{T} & = \frac{1}{B^2} \left( \left \| \mathbf{T}^{\tilde{f}}\left ( \mathbf{T}^{fc} \right)^{t}  \right \|_{F} + \left \| \mathbf{T}^{c}\left ( \mathbf{T}^{fc} \right)^{t} \right \|_{F} \right. \\
& \left. +\left \| \mathbf{T}^{c}\left ( \mathbf{T}^{cf} \right)^{t}  \right \|_{F} + \left \| \mathbf{T}^{\tilde{f}}\left ( \mathbf{T}^{cf} \right)^{t}  \right \|_{F} \right) 
\end{split}
\end{equation}
where $\left \| \cdot \right \|_{F}$ denotes the Frobenius norm. In the same way, we can get $l_{e}^{I}$ and $l_{e}^{M}$.

In summary, the overall interaction constraint loss is defined as
\begin{equation}
l_{ic}=l_{c}^{T}+l_{c}^{I}+l_{c}^{M}+l_{e}^{T}+l_{e}^{I}+l_{e}^{M}
\end{equation}

\subsection{Multi-granularity Constraints}
\label{subsubsec:Multi-granularity Feature Contrast Constraint}

\subsubsection{News Overall Constraint}
\label{subsubsec:Inter-class Constraint of news}

The news overall constraint aims to amplify the difference between real and fake news at the news level. Since the Maximum Mean Difference (MMD) can measure the gap between two distributions and has achieved promising results in domain migration \cite{liu2023robust}, we utilize MMD to accentuate the distinction between real and fake news. To elaborate, given two distributions $\mathbf{X}$ and $\mathbf{Y}$, the MMD can be formalized as,
\begin{equation}
\label{eq14}
\mathrm{MMD}^2 \left(\mathcal{F}, \mathbf{X}, \mathbf{Y} \right) = \left \| \frac{1}{n}\sum_{i=1}^{n}f(x_i) - \frac{1}{m}\sum_{j=1}^{m}f(y_j)   \right \|^2_{\mathcal{H}} 
\end{equation}
where $\mathcal{F}$ denotes the function domain. $x_i$ and $y_j$ denote the original data of the two distributions. $n$ and $m$ denote the number of the original data. $f(x_i)$ and $f(y_j)$ denote the values after mapping to the reproducing Hilbert space $\mathcal{H}$ by the kernel function $f(\cdot)$. For convenience, we utilize the Gaussian kernel function $K(\cdot)$, thus Eq. \eqref{eq14} is derived as

\begin{equation}
\label{eq15}
\begin{split}
\mathrm{MMD}^2 \left(\mathcal{F}, \mathbf{X}, \mathbf{Y} \right) &= \left \| \frac{1}{n^2}\sum_{i=1}^{n}\sum_{i' =1}^{n}K(x_i,x_{i'}) \right.\\
& \left. - \frac{2}{nm}\sum_{i=1}^{n}\sum_{j =1}^{m}K(x_i,y_j) \right. \\
& \left. + \frac{1}{m^2}\sum_{j=1}^{m}\sum_{j' =1}^{m}K(y_j,y_{j'}) \right \|
\end{split}
\end{equation}
Based on the above, we divide $\mathbf{{M}'}$ into $\mathbf{{M}'}^+$ and $\mathbf{{M}'}^-$ according to the labels (real or fake), then feed them into Eq. \eqref{eq15}. Thus, we can obtain the news overall constraint loss,
\begin{equation}
l_{no}=-\mathrm{MMD}^2 \left(\mathcal{F}, \mathbf{{M}'}^+, \mathbf{{M}'}^- \right)
\end{equation}

\subsubsection{News Internal Constraint}
\label{subsubsec:Intra-class Constraint of news}

The news internal constraint aims to amplify the difference at the feature level. For real news, it aims to minimize the divergence between the same image-text pairs, which concurrently maximizes the distance between disparate pairs. For fake news, it aims to widen the distance between all image-text pairs. This approach ensures that, for real news, the features within the same image-text pair remain closely aligned, insulated from the influence of high similarity features from other real news. Meanwhile, for fake news, it effectively severs the cohesion of all image-text pairs and keeps them all away from each other. A straightforward yet effective method to implement this is to utilize cosine similarity. To elaborate, given two distributions $\mathbf{X}$ and $\mathbf{Y}$, the cosine similarity can be calculated by
\begin{equation}
\mathrm{sim} \left ( \mathbf{X},\mathbf{Y} \right ) = \frac{\mathbf{X}\mathbf{Y}}{\left \| \mathbf{X} \right \|\left \| \mathbf{Y} \right \| }
\end{equation}
Based on the above, we feed the text features $\mathbf{{T}'}$ and image features $\mathbf{{I}'}$ as the input. Then, we divide them into $\mathbf{{T}'}^+$ and $\mathbf{{T}'}^-$, $\mathbf{{I}'}^+$ and $\mathbf{{I}'}^-$, respectively, according to the labels (real or fake). Next, for real news features $\mathbf{{T}'}^+$ and $\mathbf{{I}'}^+$, we have
\begin{equation}
\begin{split}
l_t = & - \frac{1}{B^+} \sum_{i=1}^{B^+}\left ( \mathrm{sim} \left ( \mathbf{{T}'}^+_{i}, \mathbf{{I}'}^+_{i} \right)  \right )  \\
& + \frac{1}{B^+\left(B^+-1\right)} \sum_{i=1}^{B^+}\sum_{\substack{j=1 \\ j \ne i}}^{B^+}\left ( \mathrm{sim} \left ( \mathbf{{T}'}^+_{i}, \mathbf{{I}'}^+_{j} \right) \right)
\end{split}
\end{equation}
For fake news features $\mathbf{{T}'}^-$ and $\mathbf{{I}'}^-$, we have 
\begin{equation}
l_f = \frac{1}{(B^-)^2} \sum_{i=1}^{B^-}\sum_{\substack{j=1 }}^{B^-}\left ( \mathrm{sim} \left ( \mathbf{{T}'}^-_{i}, \mathbf{{I}'}^-_{j} \right) \right ) 
\end{equation}
where $B^+$ and $B^-$ denote the number of real and fake news, respectively. Besides, they satisfy $B^+ + B^- = B$.

In this way, we can obtain the news internal constraint loss,
\begin{equation}
l_{ni} = l_t +l_f
\end{equation}

\subsection{Dominant Feature Fusion Reasoning}
\label{subsubsec:Dominant Feature Fusion Reasoning}

To fully leverage the advantages of features extracted from different modalities, we propose a dominant feature fusion reasoning module. First, we 
combine the unimodal features by $\hat{\mathbf{{M}}} = \mathbf{{T}'} + \mathbf{{I}'}$. 
Second, we design the inconsistency and consistency mining strategy to highlight the discrepancy and similarity between $\mathbf{{M}'}$ and $\hat{\mathbf{{M}}}$, respectively. For the former, we perform both subtraction and multiplication,
\begin{equation}
\mathbf{{F}^1} = \left | \mathbf{{M}'} - \hat{\mathbf{{M}}} \right |
\end{equation}
\begin{equation}
\mathbf{{F}^2} =  \mathbf{{M}'} \times \hat{\mathbf{{M}}}
\end{equation}
For the latter, we perform both concatenation and the iAFF fusion,
\begin{equation}
\mathbf{{F}^3} =  \mathrm{ProjectionHead}\left(\mathrm{Concat} \left ( \mathbf{{M}'}, \hat{\mathbf{{M}}}  \right )\right )
\end{equation}
\begin{equation}
\mathbf{{F}^4} =  \mathrm{iAFF} \left ( \mathbf{{M}'}, \hat{\mathbf{{M}}}  \right )
\end{equation}

Then, we concatenate the features extracted by the above and feed them into a classifier that consists of a two-layer fully connected network, to predict the label $\hat{y}$,
\begin{equation}
\hat{y}  =  \mathrm{Classifier} \left ( \mathrm{Concat}\left ( \mathbf{{F}^1}, \mathbf{{F}^2}, \mathbf{{F}^3}, \mathbf{{F}^4} \right )   \right ) 
\end{equation}
Here, the classifier is trained with cross-entropy loss against the ground-truth label $y$, 
\begin{equation}
l_{ce} = -\sum y\mathrm{log} \left ( \hat{y} \right) 
\end{equation}

The final loss function is as follows,
\begin{equation}
\label{eq16}
L = \lambda l_{ic} + l_{no} + l_{ni} + l_{ce}
\end{equation}
where $\lambda$ controls the ratio of $l_{ic}$.
\section{Experiments}
\label{sec:Experiments}

\subsection{Experimental Settings}
\label{subsec:Experimental Settings}
\subsubsection{Datasets}
\label{subsubsec:Datasets}
Our method is evaluated on three real-world datasets: Weibo17 \cite{jin2017multimodal}, Politifact and GossipCop \cite{shu2020fakenewsnet}, which are all widely recognized and widespread datasets in fake news detection. 
We remove unimodal news items that lack either an image or text. If the text of a news item contains multiple related images, we select the first image to match it. After preprocessing, the details of these datasets are reported in Table \ref{tab0}.
\begin{table}[!t]
	\centering
	\begin{tabular}{ccccccc}
		\hline\hline
		& \multicolumn{3}{c}{Train} & \multicolumn{3}{c}{Test} \\ \cline{2-7}
		\multicolumn{1}{c}{\multirow{-2}{*}{Dataset}} & Real   & Fake   & Total   & Real   & Fake   & Total  \\ \hline
		Weibo17    & 2800   & 3335   & 6135    & 833    & 852    & 1685   \\ 
		Politifact & 197    & 117    & 314     & 63     & 27     & 90     \\
		GossipCop  & 6796   & 1713   & 8509    & 1946   & 462    & 2408  \\ \hline\hline
	\end{tabular}
	\caption{Statistics of three fake news datasets.} %by underlining
	
	\label{tab0}
\end{table}

\subsubsection{Implementation Details}
\label{subsubsec:Implementation Details}

The proposed method is implemented with PyTorch and all experiments are performed on the NVIDIA RTX 3090 GPU.
We pad the sequence lengths of news to 300 for all datasets and resize all images to $224 \times 224$. 
% Second, we utilize the pre-trained BERT model to extract text features.
The pre-trained CLIP model is ``ViT-B/32'', where we utilize Google Translation API\footnote{http://translate.google.com} and summary generation model\footnote{https://huggingface.co/t5-large} to obtain English summaries for text longer than 50 characters.
Thus, we can get $n_1 = 768$, $N=512$, $n_2 = 1000$. 
During training, we freeze the BERT and CLIP models and fine-tune ResNet-101 for better results. 
The dropout in all projection head is set to 0.4. In multi-head attention, the attention heads and dropout are set to 8 and 0.1, respectively. 
The $\lambda$ in Eq. \eqref{eq16} is set to 0.1. Additionally, the initial learning rate and batch size are set to 0.001 and 64, respectively. We utilize the Adam optimizer for all three datasets. Finally, we trained our model for 80 epochs and chose the epoch with the best testing accuracy to avoid overfitting.

\subsection{Performance Evaluation}
\label{subsec:Performance Evaluation}
In this section, we conduct a comparison between our proposed RaCMC method with seven state-of-the-art methods on three datasets. Table \ref{tab1} displays the performance comparison results. It can be observed that RaCMC achieves the highest accuracy of 91.5\%, 92.2\%, and 87.9\% on the three datasets, respectively. Besides, compared to other methods, RaCMC always ranks 1st or 2nd on Recall and F1-score for fake news, and on Precision and F1-score for real news, across all datasets. 
Specifically, in the Weibo17 dataset, compared with seven state-of-the-art methods, 
RaCMC always ranks 1st across seven metrics, which improves the detection accuracy by 1.0\%$ \sim $15.3\%. 
In the Politifact dataset, RaCMC ranks 1st five times and 3rd twice, which improves the detection accuracy by 1.8$ \sim $18.1\%. 
In the GossipCop dataset, the performance improvement of RaCMC is small, as nearly 50\% of the images are celebrity faces that are even hard to distinguish for many human viewers \cite{ying2023bootstrapping}.
Although FND-CLIP has a very close performance with RaCMC on GossipCop, RaCMC outperforms FND-CLIP by a more noticeable 1.0\% and 2.2\% on the Weibo17 and Politifact dataset, respectively.

\begin{figure}[t]
	\centering
	\subfloat[FND-CLIP]{
		\includegraphics[width=0.23\textwidth]{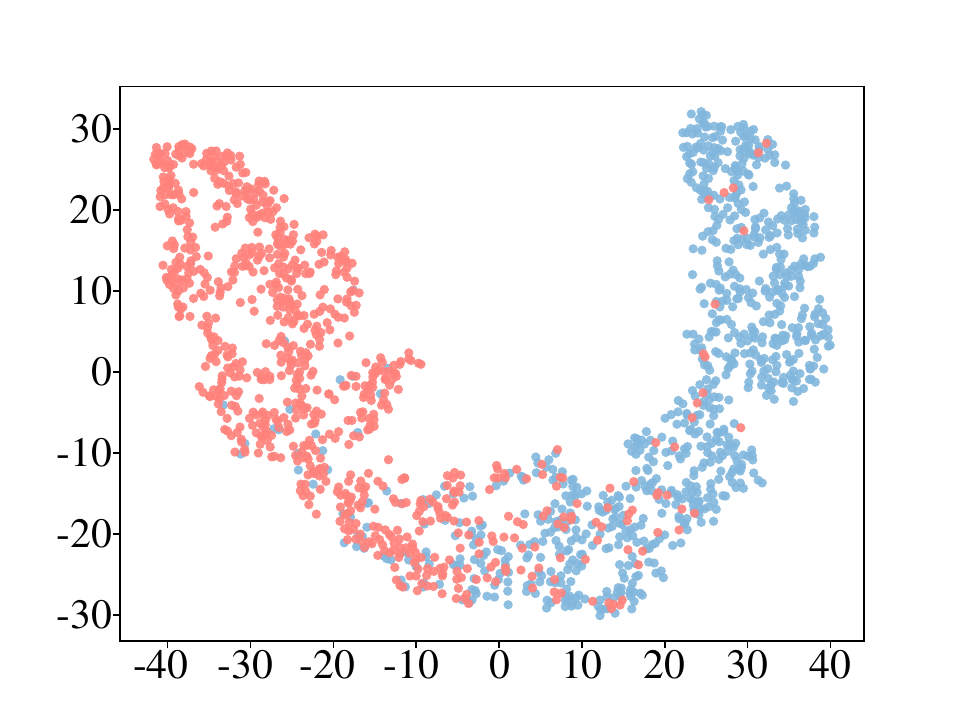}
		\label{fig3:subfig1}
	}
	\subfloat[Our RaCMC]{
		\includegraphics[width=0.23\textwidth]{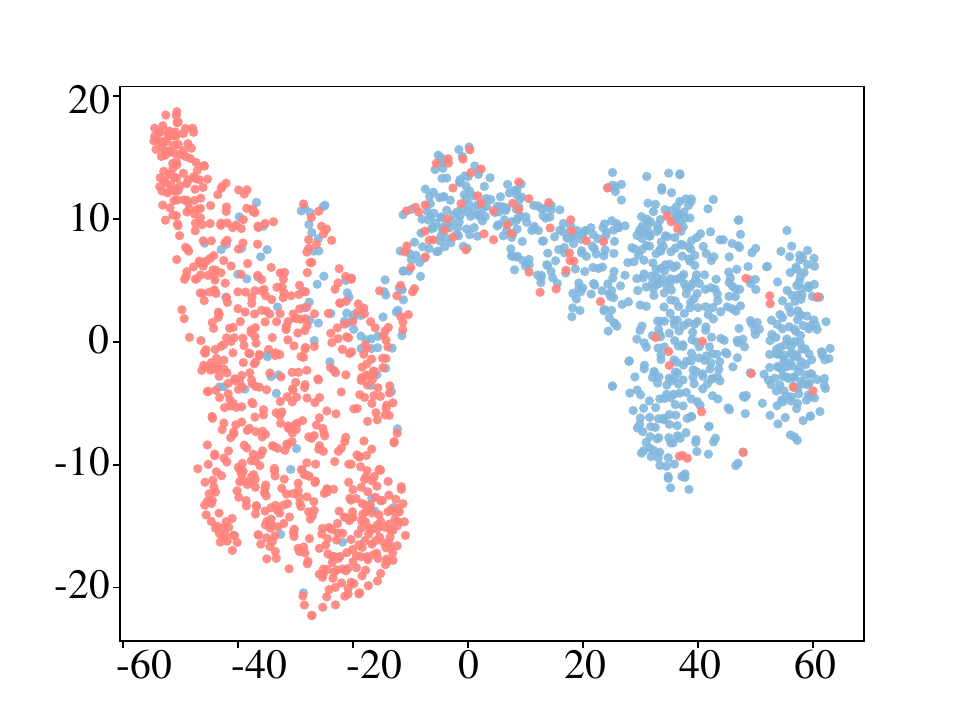}
		\label{fig3:subfig2}
	}
	\caption{t-SNE visualization of mined features on the test set of the Weibo17 dataset. Dots with the same color are within the same label.}
	\label{fig3}
\end{figure}

Moreover, we present the t-SNE visualizations of features in Figure \ref{fig3}. 
In RaCMC, the separation between the purple and pink dots is more pronounced, with a clearer boundary and a more symmetrical and concise overall structure, indicating that it extracts more distinguishable features.

\begin{table*}[t]
	\centering
	\begin{tabular}{ccccccccc}
		\hline\hline
		\multicolumn{1}{l}{}          &                        &                                        
		& \multicolumn{3}{c}{Fake News}              & \multicolumn{3}{c}{Real News}                    \\ \cline{4-9} 
		\multicolumn{1}{c}{\multirow{-2}{*}{Dataset}} & \multirow{-2}{*}{Method} & Accuracy                               & Pre.      & Rec.         & F1.       & Pre.      & Rec.         & F1.       \\ \hline
		\multicolumn{1}{l}{} & MVAE \cite{khattar2019mvae}                                            & 0.824                                  & 0.854          & 0.769          & 0.809          & 0.802          & 0.875          & 0.837          \\
		\multicolumn{1}{l}{} & SAFE \cite{zhou2020similarity}                                           & 0.762                                  & 0.831          & 0.724          & 0.774          & 0.695          & 0.811          & 0.748          \\
		\multicolumn{1}{l}{} & CAFE \cite{chen2022cross}                                           & 0.840                                  & 0.855          & 0.830          & 0.842          & 0.825          & 0.851          & 0.837          \\
		\multicolumn{1}{l}{} & FND-CLIP* \cite{zhou2023multimodal}                                       & {\color[HTML]{3531FF} \textbf{0.905}}                            & 0.893          & {\color[HTML]{3531FF} \textbf{0.921}}          &{\color[HTML]{3531FF} \textbf{0.907}} & {\color[HTML]{3531FF} \textbf{0.917}}          & 0.887    & {\color[HTML]{3531FF} \textbf{0.902}}          \\
		\multicolumn{1}{l}{} & MPFN \cite{jing2023multimodal}                                           & 0.838                                  & 0.857          & 0.894          & 0.889          & 0.873          & 0.863          & 0.876          \\
		\multicolumn{1}{l}{} & MRAN \cite{yang2024mran}                                           & 0.903                                  & {\color[HTML]{3531FF} \textbf{0.904}}          & 0.908          & 0.906          & 0.897          & 0.892          & 0.894          \\
		\multicolumn{1}{l}{} & CSFND \cite{peng2024not}                                           & 0.895                                  & 0.899          & 0.895          & 0.897          & 0.892          & {\color[HTML]{3531FF} \textbf{0.896}}          & 0.894          \\
		\multirow{-8}{*}{Weibo17}    & \textbf{RaCMC (Proposed)}                                              & \cellcolor[HTML]{FFFFFF}{\color[HTML]{FE0000} \textbf{0.915}} & {\color[HTML]{FE0000} \textbf{0.910}}    & {\color[HTML]{FE0000} \textbf{0.924}}    & {\color[HTML]{FE0000} \textbf{0.917}}          & {\color[HTML]{FE0000} \textbf{0.921}} & {\color[HTML]{FE0000} \textbf{0.906}}          & {\color[HTML]{FE0000} \textbf{0.914}}    \\ \hline
		& dEFEND \cite{shu2019defend}                  & {\color[HTML]{3531FF} \textbf{0.904}}                            & -             & -             & -             & 0.902          & {\color[HTML]{FE0000} \textbf{0.956}} & 0.928          \\
		& LSTM-ATT \cite{lin2019detecting}               & 0.832                                  & -             & -             & -             & 0.836          & 0.832          & 0.829          \\
		& Spotfake+ \cite{singhal2020spotfake+}                                      & 0.846                                  & -             & -             & -             & -             & -             & -             \\
		& GNN \cite{han2020graph}                  & 0.803                                  & -             & -             & -             & 0.806          & 0.801          & 0.801          \\
		& DistilBert \cite{allein2021like}            & 0.741                                  & {\color[HTML]{FE0000} \textbf{0.875}}    & 0.636          & 0.737          & 0.647          & 0.88           & 0.746          \\
		& CAFE \cite{chen2022cross}                                           & 0.864                                  & 0.724          & 0.778          & 0.750          & 0.895          & 0.919          & 0.907          \\
		& FND-CLIP* \cite{zhou2023multimodal}                                       & 0.900                                  & {\color[HTML]{3531FF} \textbf{0.846}}          & {\color[HTML]{3531FF} \textbf{0.815}}    & {\color[HTML]{3531FF} \textbf{0.830}}          & {\color[HTML]{3531FF} \textbf{0.922}} & {\color[HTML]{3531FF} \textbf{0.937}}          & {\color[HTML]{3531FF} \textbf{0.929}}    \\
		\multirow{-8}{*}{Politifact} & \textbf{RaCMC (Proposed)}                                              & {\color[HTML]{FE0000} \textbf{0.922}} & 0.833 & {\color[HTML]{FE0000} \textbf{0.926}}          & {\color[HTML]{FE0000} \textbf{0.877}} & {\color[HTML]{FE0000} \textbf{0.967}}    & 0.921    & {\color[HTML]{FE0000} \textbf{0.943}} \\ \hline
		& dEFEND \cite{shu2019defend}                                        & 0.808                                  & -             & -             & -             & 0.729          & 0.782          & 0.755          \\
		& LSTM-ATT \cite{lin2019detecting}                                        & 0.842                                  & -             & -             & -             & 0.839          & 0.842          & 0.821          \\
		& Spotfake+ \cite{singhal2020spotfake+}                                     & 0.858                                  & 0.732             & 0.372             & 0.494             & 0.866             & {\color[HTML]{FE0000} \textbf{0.962}}             & 0.914             \\
		& GNN \cite{han2020graph}                                          & 0.841                                  & -             & -             & -             & 0.820          & 0.831          & 0.825          \\
		& DistilBert \cite{allein2021like}                                     & 0.857                                  & {\color[HTML]{FE0000} \textbf{0.805}} & {\color[HTML]{3531FF} \textbf{0.527}}          & {\color[HTML]{3531FF} \textbf{0.637}}          & 0.866          & {\color[HTML]{3531FF} \textbf{0.960}}    & 0.911          \\
		& CAFE \cite{chen2022cross}                                          & 
		0.867                                 & 0.732          & 0.490          & 0.587          & 0.887          & 0.957          &0.921    \\
		& FND-CLIP* \cite{zhou2023multimodal}                                      & {\color[HTML]{3531FF} \textbf{0.876}} & {\color[HTML]{3531FF} \textbf{0.761}}          & 0.517 & 0.616          & {\color[HTML]{3531FF} \textbf{0.894}} & 
		{\color[HTML]{FE0000} \textbf{0.962}}          & {\color[HTML]{3531FF} \textbf{0.926}} \\
		\multirow{-8}{*}{GossipCop}  & \textbf{RaCMC (Proposed)}                                             & {\color[HTML]{FE0000} \textbf{0.879}}                            & 0.745          & {\color[HTML]{FE0000} \textbf{0.563}}          & {\color[HTML]{FE0000} \textbf{0.641}}          & {\color[HTML]{FE0000} \textbf{0.902}}    &0.954 & {\color[HTML]{FE0000} \textbf{0.927}} \\ \hline\hline
	\end{tabular}
	\caption{Comparison between RaCMC and state-of-the-art multimodal fake news detection methods on Weibo17, Politifact and GossipCop. *: The results are reproduced by us based on the open-source code. -: The results are not available from the original paper. The best performance is highlighted in bold red and the follow-up is highlighted in bold blue.} %by underlining
	\label{tab1}
\end{table*}

\subsection{Ablation Studies}
\label{subsec:Ablation Studies}
We investigate the effects of our proposed components by defining the following variations: (1) w/o $l_{ic}$: removing interaction constraints. (2) w/o $l_{no}$: removing news overall constraint. (3) w/o $l_{ni}$: removing news internal constraint. (4) I-only: only utilizing image features to classify. (5) T-only: only utilizing text features to classify. 
(6) -w/o MRC: concatenating the features of different granularities to replace MRC. (7)  -w/o MGC: removing MGC. (8)  -w/o DFR: concatenating unimodal and multimodal features to replace DFR.

From Table \ref{tab2}, we can observe that all ablation variants perform worse than the complete RaCMC model. Specifically, when removing $l_{ic}$, the accuracy drops by 1.1\% on Politifact and 3.4\% on GossipCop. It indicates the necessity of ensuring the consistency and exclusivity of the features post-interaction. When removing $l_{no}$, the decrease in accuracy on Politifact is 3.3\% and on GossipCop is 1.6\%. 
Besides, the performance of only using unimodal features is poor. For example, I-only performs the worst, which drops the accuracy by 16.4\% on Weibo17, 11.1\% on Politifact and 4.6\% on GossipCop. 
The replacement of MRC leads to an accuracy decrease by 3.3\% on Politifact.
When removing MGC, the accuracy drops by 2.2\% on Politifact and 1.7\% on GossipCop. This indicates that MGC amplifies the differences between real and fake news and reduces misclassification. 
The replacement of DFR leads to an accuracy decrease of 5.5\% on Politifact. The results demonstrate the necessity of multi-perspective reasoning for better performance.

\begin{figure}[t]
	\centering
	\subfloat[Weibo17]{
		\includegraphics[height=0.72in, width=1.05in]{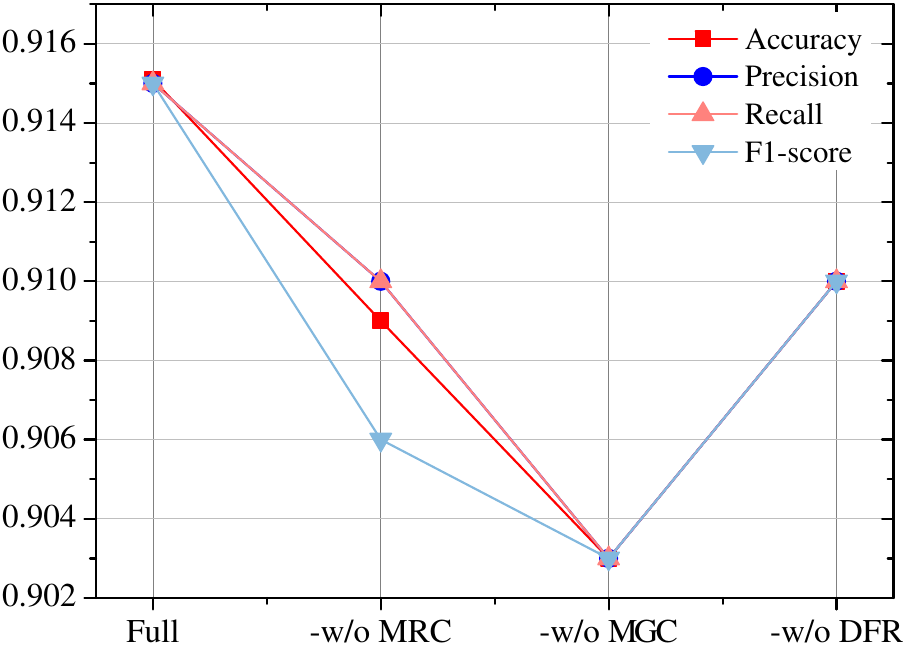}
		\label{fig4:subfig1}
	}
	\subfloat[Politifact]{
		\includegraphics[height=0.72in,width=1.05in]{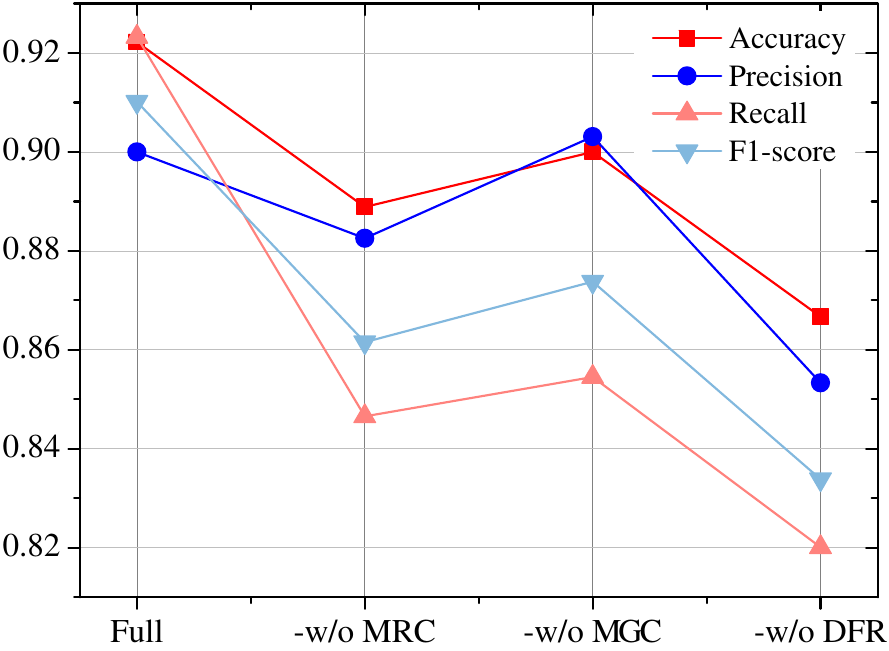}
		\label{fig4:subfig2}
	}
	\subfloat[GossipCop]{
		\includegraphics[height=0.72in, width=1.05in]{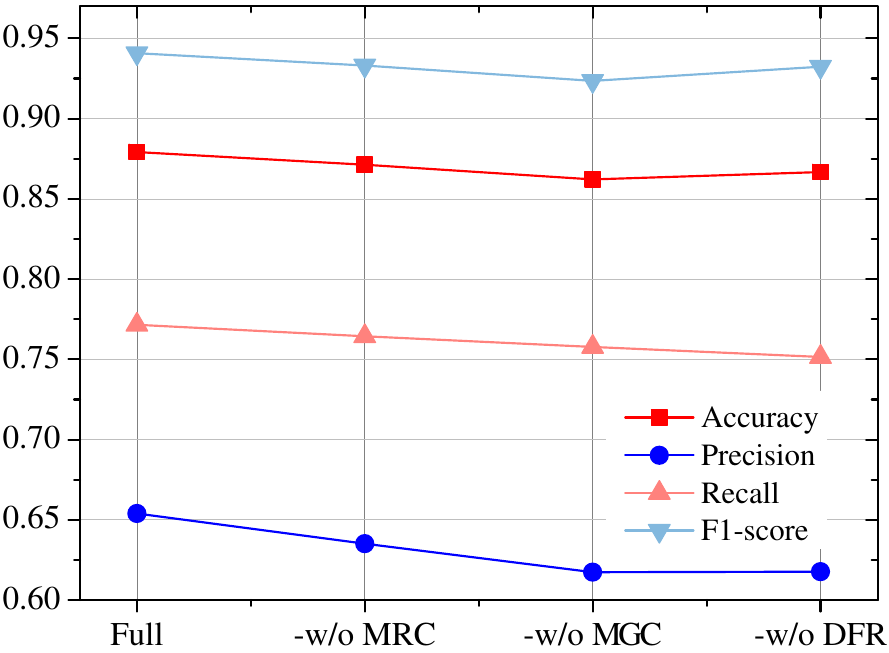}
		\label{fig4:subfig3}
	}
	\caption{Ablation study of MRC, MGC and DFR under four metrics in three datasets.}
	\label{fig4}
\end{figure}

The effects of the proposed three modules are shown in Figure \ref{fig4}, from which we can also observe that the full RaCMC always performs the best across four metrics. 

\begin{table}[t]
	\centering
	\begin{tabular}{cccc}
		\hline\hline
		\multicolumn{1}{c}{}                         & \multicolumn{3}{c}{Accuracy}                                                                                          \\ \cline{2-4} 
		\multicolumn{1}{c}{\multirow{-2}{*}{Method}}        & Weibo17                               & Politifact                            & GossipCop                             \\ \hline
		-w/o $l_{ic}$ & 0.909                                 & 0.911 & 0.845                                 \\
		-w/o $l_{no}$ & 0.903                                 & 0.889                                 & 0.863                                 \\
		-w/o $l_{ni}$ & 0.898                                 & 0.900                                 & 0.867 \\
		-I-only       & 0.751                                 & 0.811                                 & 0.833                                 \\
		-T-only       & 0.890                                 & 0.867                                 & 0.862                                 \\
		-w/o MRC     & 0.909                                 & 0.889                                 & 0.871 \\
		-w/o MGC     & 0.903                                 & 0.900                                 & 0.862                                 \\
		-w/o DFR      & 0.910 & 0.867                                 & 0.867                                 \\
		Full          & \textbf{0.915} & \textbf{0.922} & \textbf{0.879} \\ \hline\hline
	\end{tabular}
	\caption{Ablation study of RaCMC on three datasets.}
	\label{tab2}
\end{table}

\section{Conclusions}
\label{sec:Conclusions}
We propose RaCMC that sufficiently interacts and fuses cross-modal features, and amplifies the differences between real and fake news.
Residual-aware compensation interaction is proposed to acquire high-quality cross-modal multi-scale features.
Residual-aware compensation fusion is proposed to fuse and complement original information with residual blocks and multi-scale pooling.
The interaction constraints are designed to ensure the consistency and exclusivity of the features post-interaction. 
Besides, multi-granularity constraints are implemented to limit the distribution of both the news overall and the image-text pairs within the news, which amplifies the difference at the news and feature levels.
Experiments show that RaCMC performs better than a set of strong baselines.

\section{Acknowledgments}
This work is supported by the National Natural Science Foundation of China (No. 62072480, No. U2001202, No. 62172435, No. U23A20305), the National Key Research and Development Program of China (No. 2022YFB3102900), the Macau Science and Technology Development Foundation (No. SKLIOTSC-2021-2023, 0022/2022/A), the Natural Science Foundation of Guangdong Province of China (No. EF2023-00116-FST).

\bibliography{reference}

\end{document}